\documentclass{article}

\usepackage{microtype}
\usepackage{graphicx}
\usepackage{subfigure}
\usepackage{booktabs} % for professional tables

\usepackage{dblfloatfix}
\usepackage{hyperref}

\usepackage[accepted]{icml2024}

\usepackage{amsmath}
\usepackage{amssymb}
\usepackage{mathtools}
\usepackage{amsthm}

\usepackage[capitalize,noabbrev]{cleveref}

\theoremstyle{plain}

\theoremstyle{definition}

\theoremstyle{remark}

\usepackage[textsize=tiny]{todonotes}

\icmltitlerunning{Crown, Frame, Reverse: Layer-Wise Scaling Variants for LLM Pre-Training}

\begin{document}

\twocolumn[
\icmltitle{Crown, Frame, Reverse: Layer-Wise Scaling Variants for LLM Pre-Training}

% It is OKAY to include author information, even for blind
% submissions: the style file will automatically remove it for you
% unless you've provided the [accepted] option to the icml2024
% package.

% List of affiliations: The first argument should be a (short)
% identifier you will use later to specify author affiliations
% Academic affiliations should list Department, University, City, Region, Country
% Industry affiliations should list Company, City, Region, Country

% You can specify symbols, otherwise they are numbered in order.
% Ideally, you should not use this facility. Affiliations will be numbered
% in order of appearance and this is the preferred way.
\icmlsetsymbol{equal}{*}
% AUTHORS
\begin{icmlauthorlist}
\icmlauthor{Andrei Baroian}{liacs}
\icmlauthor{Kasper Notebomer}{liacs}
\end{icmlauthorlist}

\icmlaffiliation{liacs}{LIACS, Leiden University, The Netherlands}
\icmlcorrespondingauthor{Andrei Baroian}{m.a.baroian@umail.leidenuniv.nl}

%\icmlcorrespondingauthor{Firstname2 Lastname2}{first2.last2@www.uk}

% You may provide any keywords that you
% find helpful for describing your paper; these are used to populate
% the "keywords" metadata in the PDF but will not be shown in the document
\icmlkeywords{LLM, pre-training, Layer-Wise Scaling}

\vskip 0.3in
]

% this must go after the closing bracket ] following \twocolumn[ ...

% This command actually creates the footnote in the first column
% listing the affiliations and the copyright notice.
% The command takes one argument, which is text to display at the start of the footnote.
% The \icmlEqualContribution command is standard text for equal contribution.
% Remove it (just {}) if you do not need this facility.

%\printAffiliationsAndNotice{}  % leave blank if no need to mention equal contribution
%\printAffiliationsAndNotice{\icmlEqualContribution} % otherwise use the standard text.

\begin{abstract}

Transformer-based language models traditionally use uniform (isotropic) layer sizes, yet they ignore the diverse functional roles that different depths can play and their computational capacity needs. Building on Layer-Wise Scaling (LWS) and pruning literature, we introduce three new LWS variants - Framed, Reverse, and Crown - that redistribute FFN widths and attention heads via two or three-point linear interpolation in the pre-training stage.% Complementing this, we apply Grouped Query Attention (GQA) to selectively compress the KV cache.
We present the first systematic ablation of LWS and its variants, on a fixed budget of 180M parameters, trained on 5B tokens. All models converge to similar losses and achieve better performance compared to an equal-cost isotropic baseline, without a substantial decrease in training throughput. This work represents an initial step into the design space of layer-wise architectures for pre-training, but future work should scale experiments to orders of magnitude more tokens and parameters to fully assess their potential.
%Our study positions architectural heterogeneity as a low-risk, plug-and-play knob for practitioners seeking incremental improvements under fixed budgets.

\end{abstract}

\section{Introduction}

Transformer-based large language models (LLMs) have rapidly become foundational to progress across multiple domains, ranging from natural language understanding and reasoning to vision, robotics, and are regarded as a path towards artificial general intelligence.

Current LLM designs almost universally rely on isotropic architectures - all layers have the same hidden dimensions, number of attention heads, and feed-forward widths. Although this design simplifies implementation and scaling, such uniform parameter allocation may not reflect the most efficient use of model capacity. In practice, different layers of an LLM are known to play different functional roles, with earlier layers capturing local syntactic patterns and later layers responsible for high-level abstraction and reasoning \cite{jin2025exploringconceptdepthlarge} \cite{chen2024biggerdeeperbetterprobing} \cite{skean2025layerlayeruncoveringhidden}. Therefore, it is reasonable to hypothesize that different layers within the network may benefit from varying levels of computational capacity. A non-uniform, or heterogeneous, distribution of parameters across layers could lead to more efficient utilization of the model's capacity, enhancing performance, all within the constraints of a fixed parameter budget. 

With this motivation, \citet{mehta2024openelm} introduced Layer-Wise Scaling (LWS) in OpenELM, achieving superior performance with half the pre-training tokens compared to similar models, attributing much of this gain to LWS. It assigns fewer parameters to initial layers and linearly increases it toward deeper layers. Despite these promising claims, the specific contribution of Layer-Wise Scaling remains ambiguous due to the absence of targeted ablation studies that isolate its effects from other architectural modifications.

Another component adopted from \citet{mehta2024openelm} is Grouped Query Attention (GQA), which is tightly coupled with the LWS implementation in that work. Since GQA contributes to efficiency improvements and was used in the original LWS paper, we apply it consistently across all of our experiments.

A parallel line of research has emerged around the improvement of inference efficiency through parameter pruning techniques, which aim to reduce model size and computational cost by removing redundant weights from already trained LLMs \cite{prunnning_survey}. Initially, the pruning methods applied isotropic sparsity patterns, treating all layers uniformly \cite{wang2024modelcompressionefficientinference}. However, more recent approaches have introduced a diverse range of strategies and algorithms that analyze and quantify the relative importance of parameters throughout the network and pruning layers accordingly.
\citep{jin2025exploringconceptdepthlarge,chen2024biggerdeeperbetterprobing,skean2025layerlayeruncoveringhidden, he2024matterstransformersattentionneeded}

In this work, we first conduct controlled experiments to isolate and evaluate the effect of (vanilla) Layer-Wise Scaling. Secondly, inspired by the findings from the pruning literature regarding the importance of certain layers, we investigate whether initiating pre-training with architectures resembling the pruned models can enhance performance.
Specifically, we explore three novel architectural variations: (i) applying LWS while maintaining maximal size in the first and last layers - Framed LWS; (ii) a reverse framed LWS strategy, beginning with larger initial layers that gradually reduce in size; and (iii) a crown-shaped configuration, reflecting empirical evidence suggesting central layers are of greatest importance, used together with framing.

Our experiments are conducted on a 180M-parameter model, built upon the OLMo-core repository, and trained on a corpus of 5 billion tokens.
Although the experiments are based on a 190M parameter model, this scale is increasingly recognized as a reference setting to explore architectural efficiency under realistic computing constraints \cite{mehta2024openelm}. Small LLM models serve as reliable testbeds for evaluating generalization of architectural innovations \cite{hoffmann2022training}. Prior scaling studies have shown that techniques yielding efficiency gains in small models often extrapolate to larger scales when governed by principled design rules. Consequently, understanding the isolated effect of layer-wise scaling at this size provides evidence for its potential application in larger models, where sample efficiency translates directly into reduced training costs and environmental impact \cite{kaplan2020scaling}.

The code for our experiments is available open‐source at
\url{https://github.com/baroian/OLMo-custom}.

%!!! MoLA fine-tuning  MoE
%triangle, inverted triangle, diamond

%Vanilla LWS, Framed LWS, Reverse LWS, Crown LWS

%Triangle LWS, Framed LWS, 

In Section 2 we begin by reviewing related work on heterogeneous architectures and motivating our four LWS profiles. In Section 3, we formalize the Layer-Wise Scaling framework and detail its four variants. Section 3 also outlines the model architectures, training setup, and datasets. Section 4 presents the results, while Section 5 concludes with a discussion of limitations and future research.

\section{Related Work}

This section will explain the origin of Layer-Wise Scaling, present relevant pruning literature that inspired the new variants. It is followed by an explanation of Group Query Attention (GQA) and how it is applied in LWS.

\subsection{Layer Wise Scaling (LWS)}

Layer-Wise Scaling (LWS) is an architectural strategy for Transformer models that deviates from traditional isotropic scaling - where all layers share identical configurations \cite{kaplan2020scaling,hoffmann2022training}- by systematically varying structural parameters such as the number of attention heads $n_{\text{h}}^{(i)}$ and feed-forward hidden dimensions $d_{\text{FFN}}^{(i)}$ on a per-layer basis \cite{mehta2024openelm, mehta2021delightdeeplightweighttransformer}.
The core premise is that different layers fulfill distinct functional roles and thus require different computational capacities.  
OpenELM  \cite{mehta2024openelm} implements LWS by linearly interpolating scaling factors $n_{\text{h}}^{(i)}$ and $d_{\text{FFN}}^{(i)}$, becoming wider as it goes deeper. In other words, it linearly increases the number of parameters of each layer as it goes deeper. 
This strategic allocation of a model's parameter budget is motivated by the pursuit of optimized parameter allocation, enhanced model accuracy, and improved data efficiency as demonstrated by OpenELM, which achieves superior performance with twice fewer pre-training tokens compared to uniformly scaled models like OLMo \cite{mehta2024openelm}.

Although the claims are promising, the exact role of Layer-Wise Scaling remains unclear, as the authors have not presented targeted ablation studies to separate its impact from other architectural changes, revealing a gap in the literature. Despite thorough search efforts, we have found no other studies that applied the concept of layer-wise scaling during pre-training.

\subsection{Pruning Literature}

\citet{pan2025adaptpruneradaptivestructuralpruning} introduced Layer-Wise Adaptive Pruning which computes the relative importance of each decoder layer to decide which parameters to prune, and showed it is "extremely effective" in experiments on 32-layer LLaMA-3.1-8B model. Their pruned model does not prune the first and last layer; second layer starts at half the maximum parameters and linearly \textbf{decreases} to less than a quarter of the initial size. This suggests that allocating more computation to initial layers is better. The same conclusion is derived by \citet{huang2025determininglayerwisesparsitylarge} who took a theoretical perspective and proved that cutting out early layers leads to a "reconstruction error explosion" where the errors from pruning initial layers compound over the rest of the network. This would imply that this pruned architecture is relevant only for pruning. Nevertheless, we consider it worthwhile to investigate a variant of LWS that allocates more parameters to the early layers, with a linearly decreasing allocation toward the later parts of the network.

\citet{askari2025layerifestimatinglayerquality} introduced Influence Functions to estimate the layer quality and experiment on 32 layers Mistral 7B and Gemma 7B. They use this method in both pruning and expert allocation (in Mixture of Experts model), and find that the middle layers show the greatest contribution to performance. 
\citet{gao-etal-2025-mola} investigate the importance of layers in the context of Mixture of Experts LORA fine-tuning, and found superior performance by allocating more experts to the middle layers.

\begin{figure}[H]
    \vspace{-0.1cm}
    \centering
    \includegraphics[width=1\linewidth]{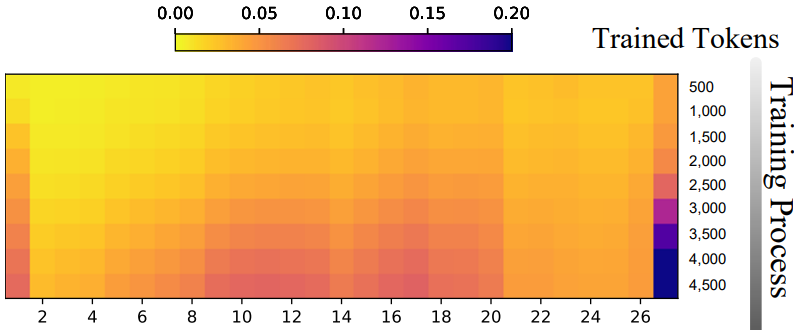}
    \caption{Figure from \citet{he2024matterstransformersattentionneeded} showing similarity score across training. Block number on x-axis, darker shades indicate higher block importance}
    \label{fig:crown_paper}
\end{figure}

The most relevant paper comes from \citet{he2024matterstransformersattentionneeded}. They \textbf{pre-train} a 7B parameter 28 layers model on 500B tokens while keeping track of the similarity score throughout the training (see Fig. \ref{fig:crown_paper}). They operate on the premise that redundant transformer blocks produce outputs highly similar to their inputs; they prune (post-training) by dropping attention layers, MLP layers, or entire transformer blocks. They concluded that deeper layers exhibit low importance and drop the number of blocks almost linearly decreasing from the initial to the final layers, but keeping the last layer almost intact. Importantly, their pruning strategy (we refer to it as Reverse LWS) does not completely reflect the findings of the similarity score (see Fig. \ref{fig:crown_paper}). In their visualization of the similarity score throughout the training, the importance of layers peaks in the middle of the network between layers 8 and 20 while the first and final layers are still of the highest importance. This inspired the Crown LWS variant.

Another insight from the same paper comes from the shifting of similarity score during pre-training, as it starts relatively flat across all layers and gradually transitions into a "crown"-shaped pattern toward the end of training.

It is important to note that in post-training compression methods, the importance of the layer is typically computed with respect to a specific dataset. \citet{tan2024dlodynamiclayeroperation} empirically showed that the contribution of individual layers varies between datasets.

Inspired by these research papers, we develop and investigate Framed LWS, Reverse LWS, and Crown LWS which will be explained in Section 3.

\subsection{Group Query Attention (GQA)}

\begin{figure}[H]
    \centering
    \includegraphics[width=0.8\linewidth]{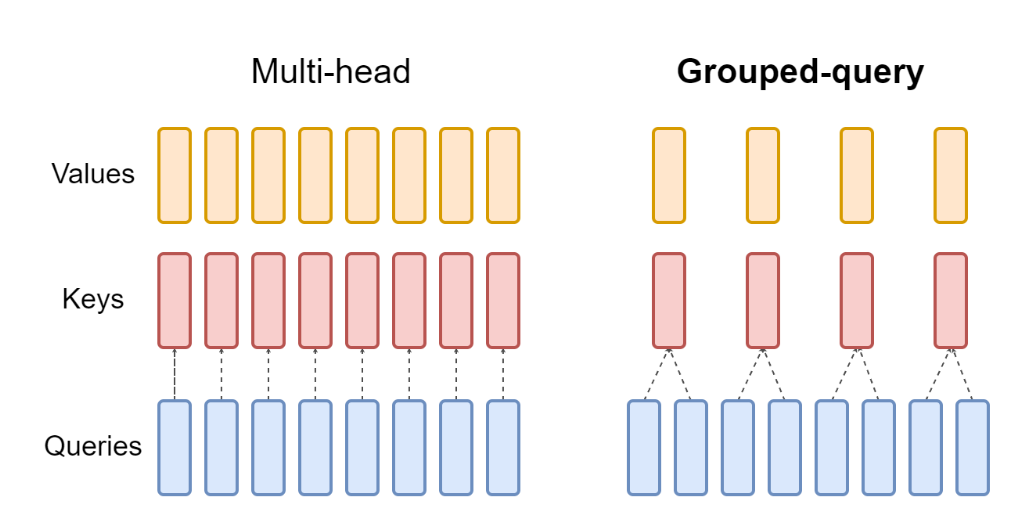}
    \caption{Grouped-Query Attention (GQA): Each group of query heads shares key and value projections, bridging MHA and MQA for improved efficiency.}
    \label{fig:models}
\end{figure}

Group Query Attention (GQA) aims to mitigate the substantial memory bandwidth and capacity demands of the Key-Value (KV) cache in standard Multi-Head Attention (MHA) during autoregressive inference \cite{ainslie2023gqatraininggeneralizedmultiquery}. GQA achieves this by dividing the total $N_q$ query heads into $G$ distinct groups, where the query heads within each group share a single Key ($K$) and Value ($V$) projection. This approach reduces the KV cache size from $N_q \times L \times d_k \times 2 \times \text{precision}$ in MHA to $G \times L \times d_k \times 2 \times \text{precision}$, where $L$ is the sequence length and $d_k$ is the key / value head dimension. Thus, GQA offers a tunable trade-off between the MHA model quality (equivalent to GQA when $G=N_q$) and the aggressive reduction of KV cache of Multi-Query Attention (MQA) \cite{shazeer2019fasttransformerdecodingwritehead} (equivalent to GQA when $G=1$), leading to improved inference efficiency while largely preserving performance.
As illustrated in Figure \ref{fig:models}, GQA groups query heads such that each group shares a single key and value projection, reducing memory overhead. This visual demonstrates how GQA interpolates between the full flexibility of MHA and the efficiency of MQA.

\textbf{Combining Layer-Wise Scaling with GQA}. The number of key/value heads \(G\) is kept \emph{fixed} in every layer.  
Scaling the attention therefore occurs solely by increasing the number of query heads \(H_q\), while the key and value heads remain \(H_k = H_v = G\).  
Because each KV-group serves exactly \(H_q/G\) query heads, the query-head count must satisfy \(H_q \bmod G = 0\). In all the experiments \(G=4\). This inhibits the impact of LWS on scaling attention as there will be fewer and rougher transitions (eg: from 8 heads to 12 heads). However, we decided to use it in all experiments, as it was utilized in the original LWS study.

\section{Methods}

In this section, the architecture of the baseline model is explained, followed by details on each of the four variants. Then the framework for building the feedforward and attention layers is presented, concluding with a description of the evaluation procedure.

\subsection{Base Model \& Training Procedure}

To evaluate the effect of Layer-Wise Scaling and its variants in isolation, we implement these changes on the OLMo 2 190M as it is part of a clean and well-maintained codebase. The complete model specifications are provided in \cite{olmo20252olmo2furious}. Worth mentioning, the model is a dense decoder-only transformer \cite{vaswani2023attentionneed} with no biases, SwiGLU activation function, Rotary positional embeddings (RoPE), RMSNorm and QK-norm. Table \ref{tab:olmo190m_hparams} shows the specific hyperparameters used for the baseline. 
Macro-batch size was chosen to fit a single GPU. Batch size (must be a multiple of macro-batch size and GPU count) and learning rate were chosen based on \citep{biderman2023pythiasuiteanalyzinglarge, brown2020languagemodelsfewshotlearners, kaplan2020scalinglawsneurallanguage} who utilize a batch size of 0.5M tokens and learning rate of 6e-4. Four KV heads create 3 GQA groups.

\begin{table}[ht]
\centering
\begin{small}
\begin{tabular}{lcc}
\toprule
\textbf{Hyper-parameter} & \textbf{Value}\\
\midrule
\addlinespace[0.3ex]
Layers ($L$) & 12 \\
Hidden size ($d_{\text{model}}$) & 768 \\
Attention heads ($H$) & 12 \\
KV heads & 4 \\
Max sequence length & 1\,024 \\
\addlinespace[0.3ex]
Macro batch size & 48 sequences \\
Global batch size & 384 sequences\\
Training steps & 1\,3000 \\
\addlinespace[0.3ex]
Learning rate & $6\times10^{-4}$ \\
Optimizer & AdamW \\
\addlinespace[0.3ex]
Tokenizer & gpt neox olmo dolma \\
Vocabulary size & 50\,279 \\
\bottomrule
\end{tabular}
\end{small}
\caption{Training hyper-parameters for Baseline 12 Layers}
\label{tab:olmo190m_hparams}
\end{table}

We train all the models on 5 billion tokens of high-quality web pages. The tokens were sampled from a version of DataComp for Language Models (DCLM) \cite{li2025datacomplmsearchgenerationtraining} filtered for quality by OLMo Team as part of DOLMino dataset mix \cite{allenai_dolmino_mix_1124}. A held-out set of 10 million tokens was created for evaluation.

\begin{figure*}[!b]
    \centering
    \includegraphics[width=1\linewidth]{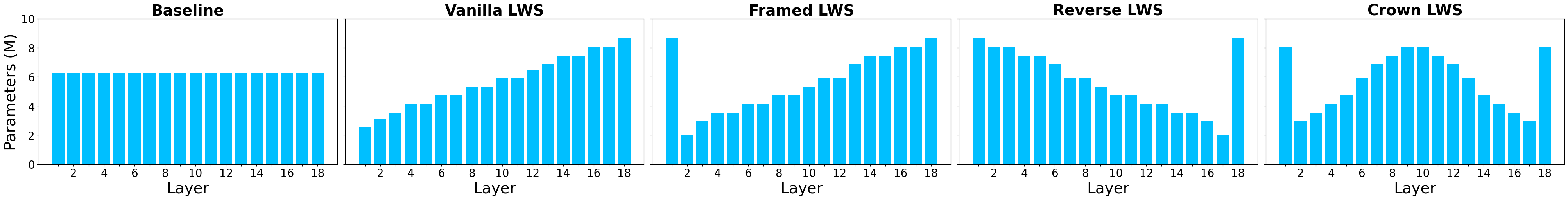}
    
    \vspace{-0.1cm}
    \caption{\textbf{Layer-Wise Scaling (LWS) variants used in our study.}  
Parameter-allocation profiles for 18 layers :  
\emph{Baseline} is an uniform (isotropic) model;
(\textit{a}) \emph{Vanilla LWS} linearly increases from shallow to deep;  
(\textit{b}) \emph{Framed LWS} keeps the first and last layers at maximum size while linearly scaling;  
(\textit{c}) \emph{Reverse LWS} more parameters allocated to initial layers, decreasing it toward later layers, using framing;  
(\textit{d}) \emph{Crown LWS} parameter count peaks in the middle layers, using framing.  
All models have the same total parameter count of 180M.}  
    \label{fig:LWS_variatns}
\end{figure*}

\subsection{Layer Wise Scaling and Variants}

%\subsection{LWS variants}

%\textcolor{red}{THE FIGURE 3 MUST BE MADE AGAIN WITH PROPER VALUES}

Using the existing literature, we test four Layer-Wise Scaling variants (see Fig. \ref{fig:LWS_variatns}):

\begin{itemize}
    \item Vanilla LWS - as used in \cite{mehta2024openelm}, linearly increasing the size of layers
    \item Framed LWS - same as Vanilla LWS but keeping the first layer at maximum dimension
    \item Reverse LWS - starting with many parameters in early layers and linearly decreasing until the end; it is framed i.e. last layer has the maximum dimension.
    \item Crown LWS - achieving maximum dimensions in the middle layers; it is framed, i.e. the first and last layers have the maximum dimension.
\end{itemize}

\textbf{Vanilla Layer-Wise Scaling} is implemented based on the methodology proposed by \cite{mehta2024openelm}. The two components of the transformer architecture that are scaled are the FNN and the attention mechanism.
The standard transformer architecture, as used in OLMo, with a given depth N, scales the FFN layer using the following formula $    d_{fnn} = \beta * d_{model}$.
%\begin{equation}
 %   d_{fnn} = \beta * d_{model}
%\label{eq:d_fnn}
%\end{equation}
Here, $d_{fnn}$ is the dimensionality of the FNN layer for the earch transformer block, $\beta$ is a fixed scalar and $d_{model}$ is the input dimension of the transformer blocks.
For layerwise scaling, this is changed to:
% \begin{equation}
%     d_{fnn}^i = \alpha^i * d_{model} \\
%     \alpha^i = 
% \label{eq:d_fnn_scaled}
% \end{equation}
\begin{align}
    d_{ffn}^i &= \beta^i * d_{model} \\
    \beta^i &= \beta_{start} + \frac{(\beta_{end}-\beta_{start}) *i}{N-1}
\label{eq:d_fnn_scaled}
\end{align}
Here $d_{fnn}^i$ is the dimensionality of the FFN for the $i$-th transformer of the $N$ total, $0 \leq i \leq N$. Every layer is scaled by $\beta^i$, which is calculated using a linear interpolation between a given $\beta_{start}$ and $\beta_{end}$.

For multi-headed attention (MHA) the default transformer has a given number of attention heads, $n_h$. The dimensionality of each head $d_h$ can then be expressed as $    d_h = \frac{d_{model}}{n_h}$.
%\begin{equation}
%    d_h = \frac{d_{model}}{n_h}
%\label{eq:head_dim}
%\end{equation}

In layer-wise scaling, the dimensionality of each attention heads is kept equal for all transformer layers; instead, the number of attention heads is scaled. The exact formulation is as follows:
\begin{align}
    n_h^i &= \alpha^i \cdot \frac{d_{model}}{d_h} \\
    \alpha^i &= \alpha_{start} + \frac{(\alpha_{end}-\alpha_{start}) *i}{N-1}
\label{eq:head_scaled}
\end{align}
Similarly to FNN, the scalar for each layer, $\alpha^i$, is a linear interpolation between a given $\alpha_{start}$ and $\alpha_{end}$. The concatenation of all the head of attention is now different in size than the dimension of the model, so the linear projection is used to scale the output of the MHA back to $d_{model}$.

%\vspace{0.1}

To achieve \textbf{Reverse LWS}, the same formulation and interpolation as LWS is used, but with the constraint that $\alpha_{start} \geq \alpha_{end}$ and $\beta_{start} \geq \beta_{end}$. This results in a parameter allocation scheme where the number of parameters per transformer decreases as the model becomes deeper.

Another variation is \textbf{Framed LWS}, where the sizes of the first and last transformers are fixed, and therefore unaffected by LWS. The propsed method achieves this by setting the scalars for the first and last transformer blocks using $\alpha = max(\alpha_{start}, \alpha_{end})$ and $\beta = max(\beta_{start}, \beta_{end})$.

The last variant is \textbf{Crown LWS}. For these models, more parameters are assigned to the middle transformer blocks compared to the first and final blocks. The same linear interpolation for $\alpha$ (\ref{eq:head_scaled}) and $\beta$ (\ref{eq:d_fnn_scaled}) is used. To achieve the crown scaling, the interpolation is now done between three scalars instead: $[\alpha_{start}, \alpha_{middle}, \alpha_{end}]$ for MHA and $[\beta_{start}, \beta_{middle}, \beta_{end}]$ for FFN. $\alpha_{start}$ and $\alpha_{end}$ represent the scalars for the first and last transformer blocks, respectively. The additional $\alpha_{middle}$ scalar controls the size of the middle transformer block.
The interpolation is performed first between $\alpha_{start}$ and $\alpha_{middle}$, and then between $\alpha_{middle}$ and $\alpha_{end}$ to calculate the scalars for all transformer blocks. The same interpolations are made for $\beta$.

%\textcolor{violet}{I'm not certain if this section is now double, but it should be fine as above are mostly implementation details, an below are exactely which ones we test i gues?}

% Reverse

% Framed

% Crown

\begin{table*}[!b]
  \vspace{-0.4cm}
  \caption{Architecture details, scaling factors, framing setting, and parameter counts.}
  \label{tab:model_specs}
  \vskip 0.15in
  \begin{center}
  \begin{small}
  \begin{sc}
  % 7 columns: l c c c c c c
  \begin{tabular}{lcccccc}
    \toprule
    Model & $n_{\text{layers}}$ & \texttt{fnn\_scalar} & \texttt{qkv\_scalar} & Framing & Params (M) & Non-embed (M) \\
    \midrule
    Baseline 12L      & 12 & [4.0, 4.0]       & [1.0, 1.0]        & false & 181.1 & 142.5 \\
    Vanilla LWS 12L       & 12 & [2.0, 5.3]       & [0.5, 2.0]        & false & 178.8 & 140.1 \\
    Baseline 18L      & 18 & [2.5, 2.5]       & [0.75, 0.75]      & false & 183.5 & 144.8 \\
    Vanilla LWS 18L       & 18 & [1.0, 4.0]       & [0.5, 1.0]        & false & 179.7 & 141.1 \\
    Framed LWS  & 18 & [0.5, 4.0]       & [0.5, 1.0]        & true  & 179.4 & 140.7 \\
    Reverse LWS   & 18 & [4.0, 0.5]       & [1.0, 0.5]        & true  & 179.4 & 140.7 \\
    Crown LWS  & 18 & [0.5, 3.8, 0.5]  & [0.5, 1.0, 0.5]   & true  & 181.9 & 143.3 \\
    \bottomrule
  \end{tabular}
  \end{sc}
  \end{small}
  \end{center}
  \vskip -0.1in
\end{table*}

\subsection{Evaluation}

The evaluation of the LWS variants will be made based on validation perplexity. This part will motivate our choice and explain why reporting results on benchmarks or generated text is infeasible.

\citet{biderman2023pythiasuiteanalyzinglarge} trained a range of model sizes from 70M to 12B parameters and share their results on common benchmarks suited for the small-scale pre-training phase. For SciQ benchmark \cite{SciQ} All the models show no growth signal until the 10 billion tokens from where the accuracy increases sharply from 0.2 to 0.6-0.8 until the 100B tokens mark, followed by a plateau or slow increase until the full 300B token training. Other benchmarks show similar growth while some display no discernible learning signal. This together with our early experiments on 4B tokens which did not showed any learning signal on any benchmarks, led us to abandon evaluating the LWS methods on benchmarks and instead focus on perplexity.

Attempts to introduce changes in architectures for language models use validation perplexity \cite{zhou2024brainformerstradingsimplicityefficiency} as well as training loss \cite{du2024stackingtransformerscloserlook}.  In their analysis of hyperparameter configurations, \citet{olmo20252olmo2furious} adopted training loss as the reference metric. Moreover, the conclusions of \cite{kaplan2020scaling} and \cite{hoffmann2022trainingcomputeoptimallargelanguage} are based on training cross-entropy loss. These give us confidence that validation perplexity is a good measurement for evaluating the LWS's performance.
 
Perplexity measures a model’s uncertainty by reporting the geometric mean of the inverse probabilities it assigns to each true next token \cite{JurafskyMartin2025perplexity}
, defined as $ \mathrm{PPL}(W) \;=\; P(w_{1:N})^{-1/N} $ and computed in our context as $\mathrm{PPL} \;=\; \exp\bigl(\mathcal{L}_{\mathrm{CE}}\bigr)$.

\section{Experiments \& Results}

The initial OLMo2-190M was chosen as the baseline, with all layers having the same dimension, but it has only 12 layers. With so few layers, the interpolation occurring in LWS would produce noticeable jumps rather than a smooth progression, especially if we are using framing where two fewer layers are scaled. As a response to this, we increased the number of layers to 18 for all models including the baseline, keeping the number of parameters the same. In other words, we prefer to exchange depth for width in our experiments, although it might affect performance as "in general, deeper models have lower perplexity". \cite{petty2024impactdepthcompositionalgeneralization}. We still run the baseline and vanilla LWS with 12 layers and compare them with the 18 layers variants.

To account for the randomness in the training process that affects the results, we had the option to either run each experiment two times or run the baseline five times giving our compute constraints. We chose the latter and ran the 18-layer baseline five times, recording the standard deviation to determine statistical significance of the results.

Table \ref{tab:model_specs} lists all the models used in our experiments including the parameter count. All variants include GQA, that is why the\textit{baseline 12L} has 180M tokens instead of the 190M from olmo2-190M configuration.

All experiments were run on Snellius \cite{snellius}, on a single node with 4 H100 GPUs, each running for approximately three hours.

\subsection*{Results}

\begin{figure}[h]
    \centering    \includegraphics[width=1.0\linewidth]{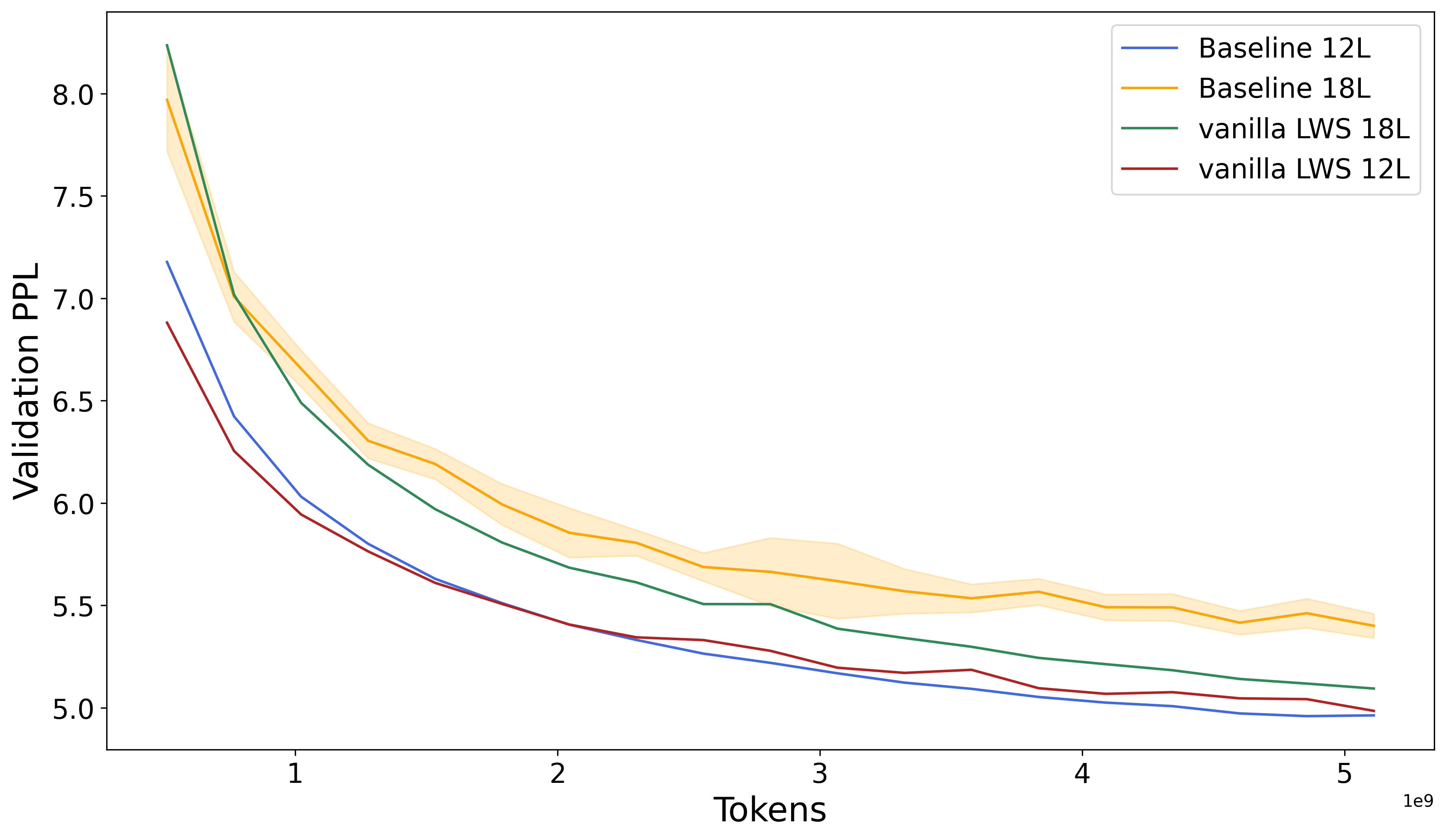}
    \vspace{-0.5cm}
    \caption{Validation Perplexity for LWS vs Baseline at 12 and 18 layers}
    \label{fig:lws_base}
\end{figure}

Table \ref{tab:val_ppl} shows the tokens per second, validation loss and validation perplexity for all models. 
First, we will look at the baseline against vanilla LWS on both 12 layers and 18 layers. It can be clearly observed in Figure \ref{fig:lws_base} that the difference is insignificant at the 12 layers, but it is meaningful at the 18 layer models, vanilla LWS having a 6\% better perplexity. The results on 12 layers are expected as the transitions from one layer to another are rough, showing that LWS works only on deeper networks. We can therefore hypothesize that on bigger models with more layers, the transition of LWS would be smoother, thus its benefits greater. 

\begin{figure}[H]
    \centering
    \vspace{-0.3cm}
    \includegraphics[width=1.0\linewidth]{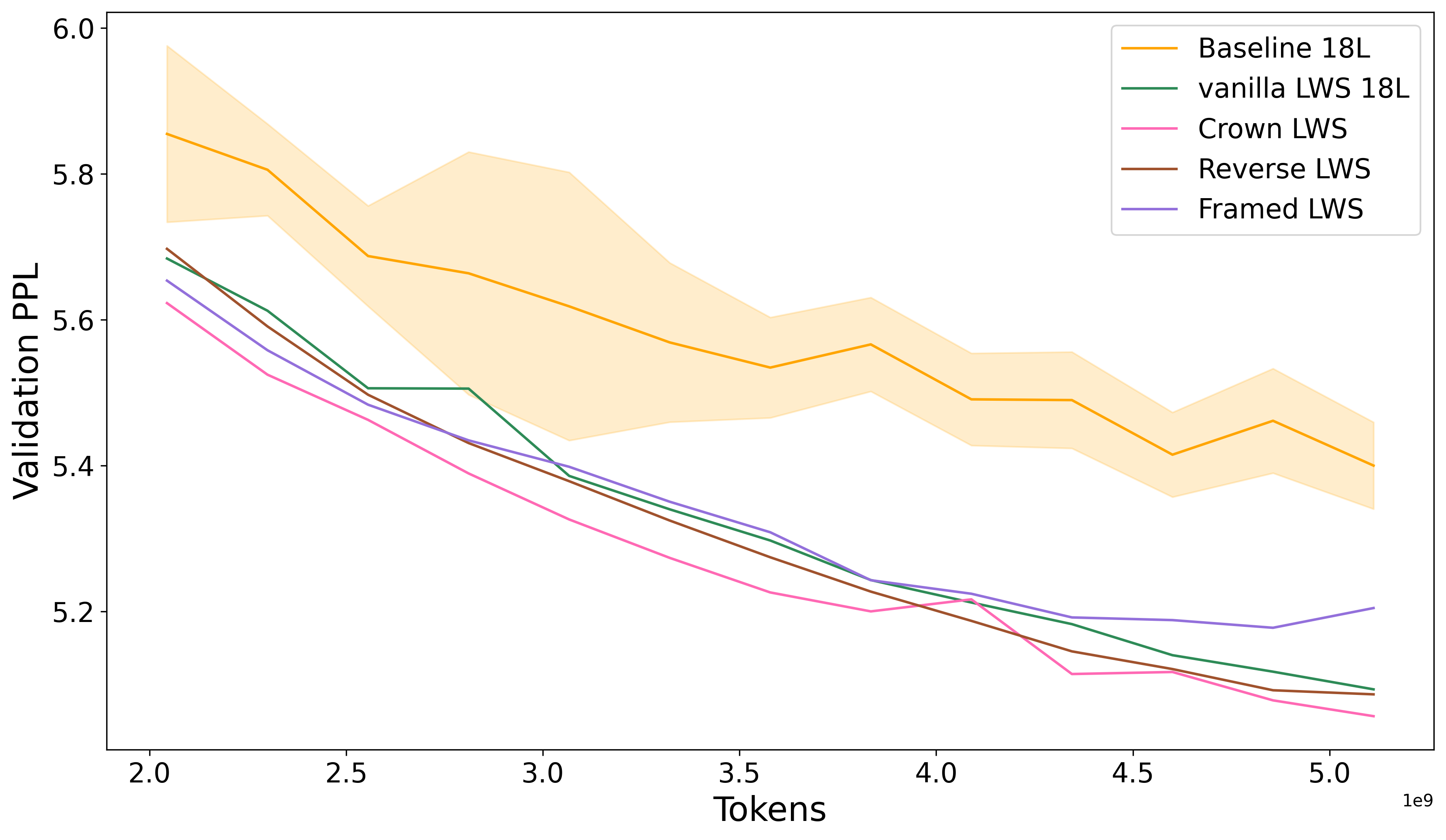}
    \vspace{-0.5cm}
    \caption{Validation Perplexity on LWS variants, all 18 Layers; \\ Zoomed in for readability by displaying results from 2B token on}
    \vspace{-0.2cm}
    \label{fig:all_18}
\end{figure}

Figure \ref{fig:all_18} shows the results of all the variants. Unexpectedly, all the variants performed almost the same, all significant better than the baseline. To exclude the possibility that this convergence arose from implementation error, for example, inadvertently training of identical architectures, we manually inspected the run logs and confirmed that the correct dimensions were instantiated in every run.
%Upon seeing the results, we double-checked if the run contained errors that might cause all the runs to have the exact model. 
Framed LWS though records a significant higher final perplexity than the other variants, but we cannot conclude that it is worst; looking at the figure, it had a bump in PPL right before the final steps, bump that can also be observed to vanilla LWS at around 2.7B tokens and Crown LWS at 4B tokens, and both of them recovered sharply. Results suggest that the presence of heterogeneity, not its exact shape, matters.

Looking at the tokens per second in table \ref{tab:val_ppl}, the LWS variants do not slow down the training process for the 18L models, but it does show a notable decrease in TPS at the 12 layers. 
The $\approx10\%$ drop in training speed training speed showed in the 12-layer models is explained as the uniform model repeats exactly the same matrix size in every feed-forward block, allowing the GPU to reuse the same low-level kernel throughout each training step.  The LWS 12L model, by contrast, introduces different layer sizes and every time the size changes the GPU has to load a different kernel, adding a setup overhead. We hypothesize that the slow-down does not appear when comparing Baseline 18L with the LWS variants as we have created the configuration for the baseline and chose scalars without taking into consideration effective use of GPUs.

Training Cross Entropy Loss can be seen in Appendix \ref{train_loss}

\begin{table}[htbp]
  \caption{Final validation perplexity (PPL) for all models after step $\ge 1000$. Standard deviation of Baseline 18L on last step is 0.0595 \\ TPS = Tokens per Second processed (i.e. how fast it trains)}
  \label{tab:val_ppl}
  \vskip 0.15in
  \begin{center}
  \begin{small}
  \begin{sc}
  \begin{tabular}{lccc}
    \toprule
    Model & TPS & Val.\ Loss & Val.\ PPL \\
    \midrule
    Baseline 12L &    \textbf{146k}  & 1.6018 & \textbf{4.962} \\
    vanilla LWS 12L  &  138k & 1.6062 & 4.984 \\
    \midrule
    Baseline 18L &  124k   &  1.6864 & 5.400 \\
    vanilla LWS 18L &  124k & 1.6279 & 5.093 \\
    Framed LWS     &  123k & 1.6490 & 5.205 \\
    Reverse LWS    &   123k & 1.6266 & 5.087 \\
    Crown LWS    &  121k  &  1.6206 & \textbf{5.057} \\
    \bottomrule
  \end{tabular}
  \end{sc}
  \end{small}
  \end{center}
  \vskip -0.1in
\end{table}

\section{Discussion}

This work represents an initial step into the design space of layer-wise heterogeneous architectures for pre-training.

All LWS variants have showed improvement up to 5\% in performance compared to the uniform baseline. This reveals that any heterogeneous allocation of parameters is preferable to a uniform one at equal compute, yet the exact profile (vanilla, framed, reverse, or crown) matters little. In other words, LWS reallocates rather than creates representational capacity, yielding incremental gains rather than transformative improvements.

This incremental benefit is far from the expectations created by \citet{mehta2024openelm}. Based on our experiments we conclude that LWS alone does not result in better or similar performance than a baseline trained on double the tokens, as claimed in their paper. The performance gains claimed in their study therefore arises from synergies between LWS and other components in the OpenELM training recipe.

Many post-training compression techniques \cite{he2024matterstransformersattentionneeded, huang2025determininglayerwisesparsitylarge, pan2025adaptpruneradaptivestructuralpruning, gao-etal-2025-mola, askari2025layerifestimatinglayerquality} have shown that transformer-based models can prune a significant number of parameters with only minor losses in downstream performance, creating efficiency gains at inference. We have questioned if we could bring those gains at the pre-training stage. A direct analogy would involve comparing a large uniform model to a slightly smaller LWS-based model and assessing whether the performance remains comparable. Instead, we chose to fix model size across both configurations, hypothesizing that Layer-Wise Scaling (LWS) could lead to improved performance at equivalent parameter count. A limitation of this setup is that it does not allow us to directly assess whether the observed efficiency improvements align quantitatively with those reported in post-training pruning studies. We nevertheless observe consistent performance gains across all LWS configurations, suggesting that even during pre-training, careful redistribution of parameters can enhance model effectiveness, resonating with the insights from pruning literature.

\subsection{Weakneseess}

The limited scale of our experiments, particularly the 5B-token training corpus, introduces uncertainty to the conclusions. Training on 100 billion tokens would have enabled evaluation on downstream benchmarks and provided a more rigorous assessment of the proposed approaches.
%Training on 100B tokens would have revealed results on downstream benchmarks and a much better measurement of the methods;
;it would also enable evaluation of the generated text, which is incoherent at our scale.

Another break point of our experiments could be the unusual low training loss as well as validation perplexity of 5 and CE loss while showing a healthy training trend - loss going gradually down with sporadic bumps. All the papers cited that pre-train LLMs have a final validation perplexity between 10 and 15, models that are 10 to 50 times bigger and trained on 60 to 500 times more tokens. The content, size and token distribution of the data set were examined, and no irregularities or sources of concern were observed.
%The data content, data size and token distribution have been checked, they do not raise any concern.
A possible reason is the data distribution as the training set comes from the same data source, a more diverse data mix would result in a higher perplexity and a higher generalization. A future investigation could test this hypothesize by using a different validation set outside of current distribution.

Another weakness comes from the compute-constraints, as running all experiments five times would give a clearer delimitation between the models performance. The LWS variants could have higher standard deviation compared to the baseline.

%UNUSUAL LOW PERPLEXITY

%Low amount of tokens

%LWS and MoE ??

\subsection{Future Work}

Our results highlight LWS's potential, but further exploration is needed to fully understand its benefits across model and dataset sizes.  At first, future work should test LWS variants on multiple, bigger models, from 190M to 7B parameters model, the size where it becomes clear if a change transfers to larger models, and to train on a much bigger corpus starting with 100B tokens and reach OLMo2-7B scale of 3B tokens.

\section{Conclusion}

Our experiments revisited Layer-Wise Scaling (LWS) for transformer language models, introducing three pruning-inspired variants (Framed, Reverse, and Crown LWS) under a fixed 180 M-parameter budget, and showed that all LWS variants improve in validation perplexity over an isotropic baseline. We found that LWS alone does not replicate the two times data efficiency gains previously claimed by OpenELM. While these findings underscore the value of heterogeneity, our study is limited by a modest 5 B-token corpus, evaluation solely via perplexity, and single-run training, suggesting that future work should involve larger-scale pre-training ($\ge 100$ B tokens, $\ge 7$ B parameters).

\section*{Acknowledgements}

We would like to express our sincere gratitude to Dr. Hazel Doughty for offering this seminar and for her support throughout the project. We also thank the Teaching Assistants, Andrius Bernatavicius and Luc Sträter, for helping to create an engaging learning environment.

We are grateful to Prof. Dr. Rob V. van Nieuwpoort for granting us access to Snellius compute resources and placing his trust in our work.

Finally, we would like to thank Dirk Groeneveld, Principal Researcher at the Allen Institute for AI, whose early advice helped us choose a focused and feasible topic, and avoid overly ambitious or saturated research directions.

\clearpage

\bibliography{drl}
\bibliographystyle{icml2024}

\newpage
\appendix
\onecolumn
\section{Training Loss}
\label{train_loss}

\begin{figure}[H]
    \centering
    \includegraphics[width=0.8\linewidth]{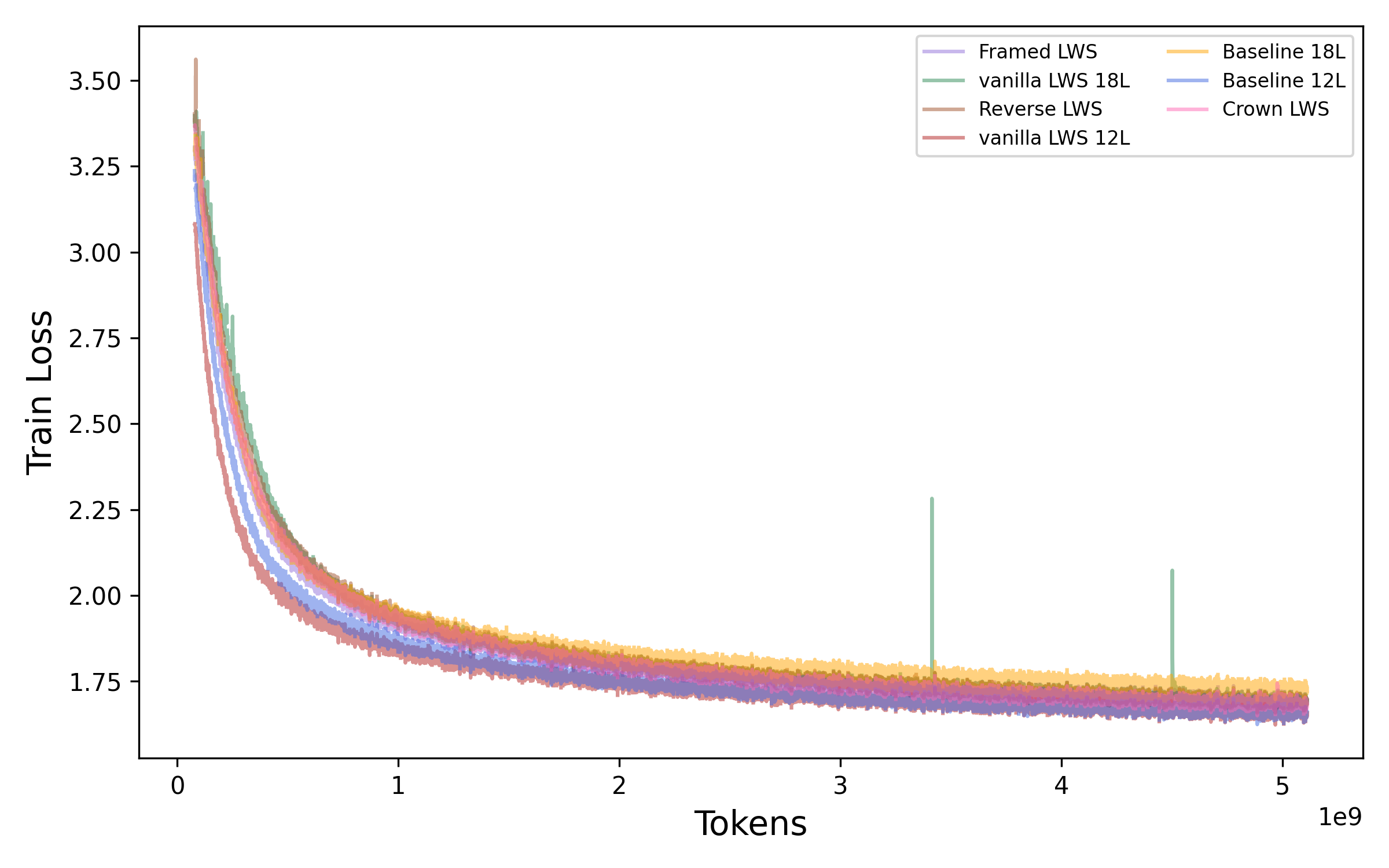}
    \caption{Training Cross Entropy Loss}
    \label{fig:train_loss}
\end{figure}

\begin{figure}[H]
    \centering
    \includegraphics[width=0.8\linewidth]{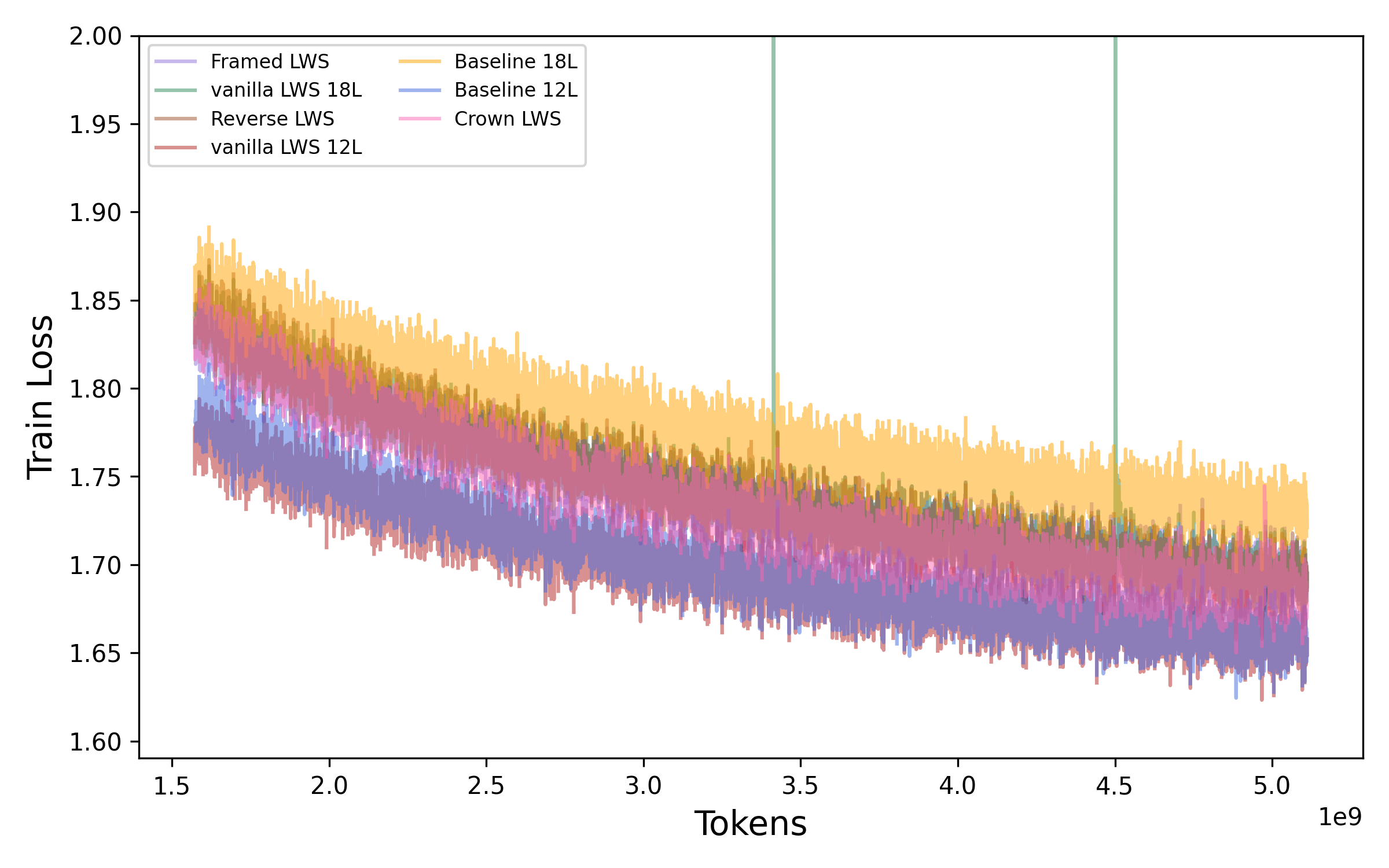}
    \caption{Training Cross Entropy Loss Zoomed in}
    \label{fig:train_loss_zoom}
\end{figure}

\end{document}